\documentclass[nohyperref]{article}

\usepackage{microtype}
\usepackage{graphicx}
\usepackage{subfigure}
\usepackage{booktabs} % for professional tables

\usepackage{hyperref}

 \usepackage[accepted]{ai4science}
 
\usepackage{amsmath}
\usepackage{amssymb}
\usepackage{mathtools}
\usepackage{amsthm}

\usepackage[capitalize,noabbrev]{cleveref}

\theoremstyle{plain}

\theoremstyle{definition}

\theoremstyle{remark}

\usepackage[textsize=tiny]{todonotes}

\icmltitlerunning{On the Relationships between Graph Neural Networks for the Simulation of Physical Systems and Classical Numerical Methods}

\begin{document}

\twocolumn[
\icmltitle{On the Relationships between Graph Neural Networks for the Simulation of Physical Systems and Classical Numerical Methods}
% On the Physical Roots and Numerical Relations of Graph Network Acceleration



\icmlsetsymbol{equal}{*}

\begin{icmlauthorlist}
\icmlauthor{Artur Toshev}{yyy,zzz}
\icmlauthor{Ludger Paehler}{yyy}
\icmlauthor{Andrea Panizza}{zzz}
\icmlauthor{Nikolaus Adams}{yyy}

\end{icmlauthorlist}

\icmlaffiliation{yyy}{Department of Mechanical Engineering, Technical University of Munich, Munich, Germany}
\icmlaffiliation{zzz}{ML Collective}

\icmlcorrespondingauthor{Artur Toshev}{artur.toshev@tum.de}

\icmlkeywords{}

\vskip 0.3in
]

\printAffiliationsAndNotice{} 

\begin{abstract}
    Recent developments in Machine Learning approaches for modelling physical systems have begun to mirror the past development of numerical methods in the computational sciences. In this survey we begin by providing an example of this with the parallels between the development trajectories of graph neural network acceleration for physical simulations and particle-based approaches. We then give an overview of simulation approaches, which have not yet found their way into state-of-the-art Machine Learning methods and hold the potential to make Machine Learning approaches more accurate and more efficient. We conclude by presenting an outlook on the potential of these approaches for making Machine Learning models for science more efficient.
\end{abstract}

\section{Introduction}

%Storyline:
%\begin{itemize}
%    \item There exist strong parallels between the recent development trajectory of machine learning approaches, and the past development of classical methods in the computational sciences
%    \item In this work we summarise existing learned solver approaches and propose ideas for further investigation.
%    \item An example of this is the recent development of GNNs mirroring the developement from Molecular Dynamics to Smoothed Particle Hydrodynamics
%    \item With ever more equivariance findings its way into modern approaches, it begs the analogy to Noether's theorem
%    \item Accelerating adoption of ML-approaches in Engineering
%    \item "biases" engineered into the machine learning approach itself make it more efficient
%    \item Summary of approaches in engineering, which present opportunities to see more ideas from engineering accelerate machine learning and make it more efficient
%    \item Categorization as: direct, single-step, multi-step autoregressively.
%\end{itemize}

%--------------------------------------------

Recent years have seen an ever-larger push towards the application of Machine Learning to problems from the physical sciences such as Molecular Dynamics~\cite{musaelian2022learning}, coarse-graining~\cite{wang2022generative}, the time-evolution of incompressible fluid flows~\cite{wang2020towards}, learning governing equations from data~\cite{brunton2016discovering,cranmer2020discovering}, large-scale transformer models for chemistry~\cite{frey2022neural}, and the acceleration of numerical simulations with machine learning techniques~\cite{kochkov2021machine}. All of these algorithms build on the infrastructure underpinning modern Machine Learning in combing state-of-the-art approaches with a deep understanding of the physical problems at hand. This begs the questions if there exist more insights and tricks hidden in existing, classical approaches in the physical sciences which have the potential to maybe not only make the algorithm for the particular problem class more efficient, but maybe even Machine Learning in general?

Inspired by recent theoretical advances in the algorithmic alignment between Graph Neural Networks (GNNs) and dynamic programming \cite{xu2020what,Velickovic2020Neural}, we surmise that the extension of this analysis to classical PDE solvers, and the physical considerations they incorporate, enables us to learn from the development trajectory in the physical sciences to inform the development of new algorithms. In this workshop paper we make the following contributions towards this goal:
\begin{itemize}
    \item A comparison of the development of graph-based learned solvers, and the proximity of their development ideas to the development of Smoothed Particle Hydrodynamics starting from Molecular Dynamics in the physical sciences.
   \item An analysis of classical numerical solvers, and their algorithmic features to inform new ideas for new algorithms.
\end{itemize}

%If the ultimate goal of ML would be constructing a general learned PDE solver capable of replacing each of these methods, then a good starting point is looking at the characteristics of each of these specialized methods and biasing the learned solver with this information. Here, we compare previous work on learned PDE solvers with their conventional solver counterparts in terms of algorithmic alignment. This section is split into commonly used tricks in learned PDE solvers and their roots in physics. 

%. Following this idea, we look at a bunch of such classical solvers and discuss their algorithmic features.

\section{MeshGraphNets and its relation to classical methods}
% Parallels in Graph Network Acceleration Development and Particle Modelling

% Introductory Figure
\begin{figure}[t]
    \centering
    \includegraphics[width=0.45\textwidth]{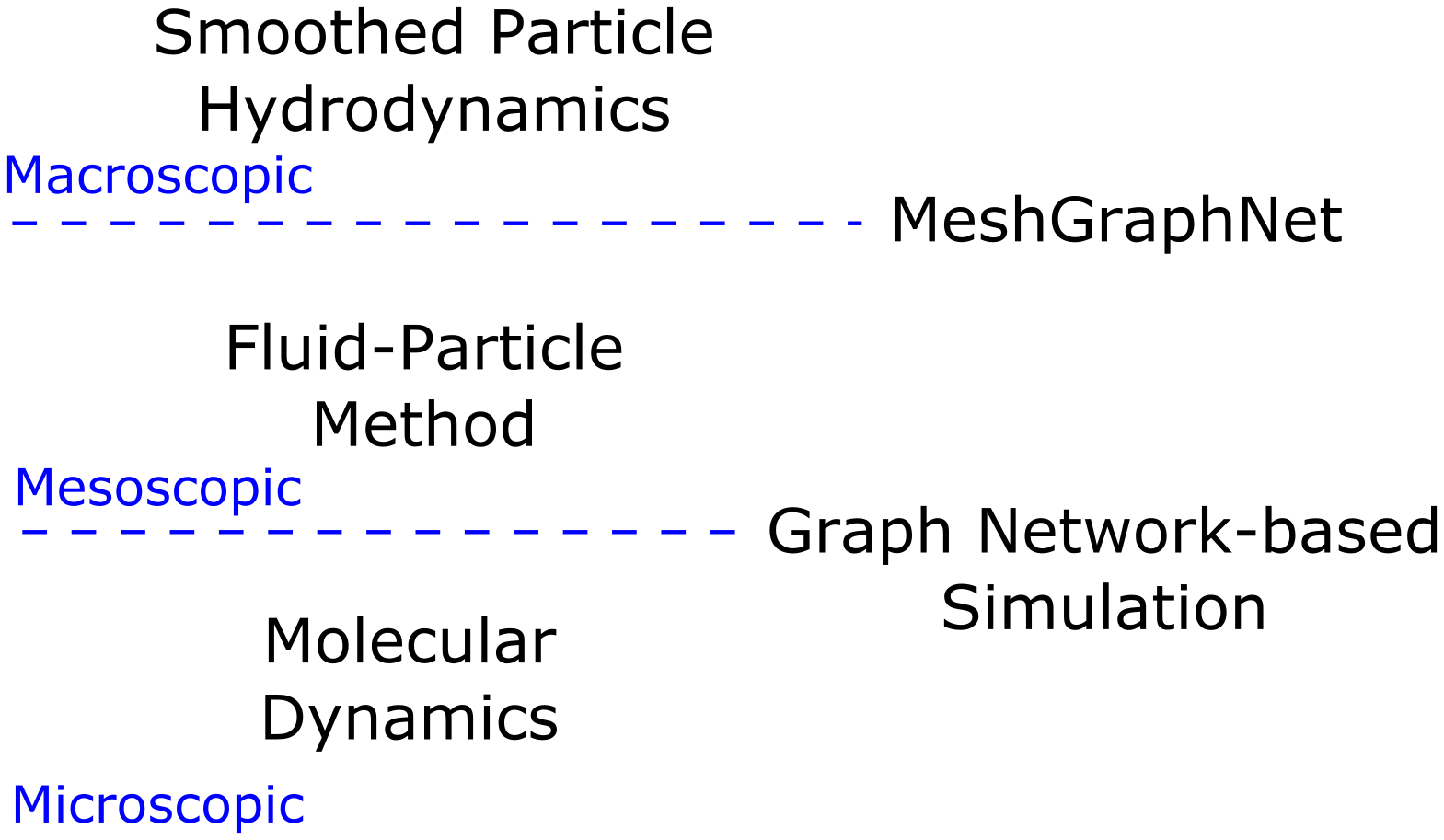}
    \caption{Characterization of the physical scales the example methods of section 2 operate on. The Graph Network-based approaches MeshGraphNets, and Graph Network-based Simulators are placed in relation to their classical counterparts.}
    \label{fig:sec2_overview}
\end{figure}

An excellent example of the parallels between the development of Machine Learning methods for the sciences and the development of classical approaches is the recent development of graph-based simulators. When we relate their inherent assumptions and techniques to the development of particle-based methods, starting with Molecular Dynamics, a great many parallels arise. For an impression of the scales the classical methods operate on, and where graph-based simulators are placed in relation, please refer to Figure~\ref{fig:sec2_overview}. 

In this section, we analyze the structure of two of the first mature learned solvers  (GNS~\cite{sanchez2020learning}, MeshGraphNets~\cite{pfaff2021learning})  and how these two approaches align with three of the classical methods (MD, FPM, SPH). We select these learned algorithms because they were one of the first of their kind to show promising results on real world data. Also, GNS is trained directly on SPH data which further motivates an algorithmic comparison.

\subsection{Graph Neural Network-based Approaches to Simulation}

The Graph Network (GN) \cite{battaglia2018relational} is a framework that generalizes graph-based learning and specifically the Graph Neural Network (GNN) architecture by \citet{scarselli2008graph}. However, in this work, we use the terms GN and GNN interchangeably. Adopting the Graph Network formulation, the main design choices are the choice of update-function, and aggregation-function. For physics-informed modeling this gives us the ability to blur the line between classical methods and graph-based methods by including biases similar to CNNs for non-regular grids, as well as encoding physical laws into our network structure with the help of spatial equivariance/invariance, local interactions, the superposition principle, and differential equations. E.g. translational equivariance can easily be incorporated using relative positions between neighboring nodes, or the superposition principle can be encoded in graphs by using the summation aggregation over the representation of forces as edge features.

Viewing MeshGraphNets~\cite{pfaff2021learning} from a physics-motivated perspective, we argue that MeshGraphNets originate from Molecular Dynamics. To present this argument in all its clarity, we have to begin with its predecessor: the Graph Network-based Simulators (GNS)~\cite{sanchez2020learning}.

\subsubsection{Graph Network-based Simulators}
The Graph Networks-based Simulator builds on the encoder-processor-decoder approach, where Graph Networks are applied iteratively on the encoded space. Proving GNS' ability to simulate systems with up to 85k particles, their approach can be summarized as follows.

Let $X^t$ denote the states of a particle system at time $t$. $X$ might contain the position, velocity, type of particle, or any other physical information specific to a material particle. A set of $k+1$ subsequent past states

\begin{equation*}
    {\bf{X}}^{t_{0:K}} = \left\{ X^{t_{0}}, X^{t_{1}}, \ldots, X^{t_{k}} \right\}
\end{equation*}

if given to the network. The core task is to then learn the differential operator $d_{\theta}$, which approximates the dynamics

\begin{equation*}
    d_{\theta}: X^{t_{k}} \longrightarrow Y^{t_{k}}, \quad X^{t_{k+1}} = \text{Update}\left\{ X^{t_{k}}, d_{\theta} \right\}.
\end{equation*}

Here, $Y^t$ is the acceleration, which is used to obtain the next state $X^{t+1}$ via integration using a deterministic "Update" routine, e.g. semi-implicit Euler scheme. The differential operator $d_{\theta}$ is learned with the encoder-processor-decoder approach where the encoder takes in 1 to 10 previous states, and encodes them into a graph. This graph consists of nodes - latent representation of the states $X$ - and edges - between each pair of particles closer than some cut-off radius there is another latent vector, which initially contains the distance or displacement information. The processor is then a multilayer Graph Network of which the exact number of message-passing Graph Networks is a hyperparameter. The result on the graph-space is then decoded back to physical space. The loss is computed as the mean-squared error between the learned acceleration, and the target acceleration. While the approach showed promising results for fluid simulations, and fluid-solid interactions, it struggled on deforming meshes, such as thin shells.

\subsubsection{MeshGraphNets}
To better represent meshes MeshGraphNets~\cite{pfaff2021learning} supplemented the Graph Network simulation with an additional set of edges to define a mesh, on which interactions can be learned. Closely related to the superposition principle in physics, the principle of splitting a complicated function into the sum of multiple simpler ones, the interaction function is split into the interaction of mesh-type edges and collision-type edges.

Following the widespread use of remeshing in engineering, MeshGraphNets have the ability to adaptively remesh to model a wider spectrum of dynamics. Mesh deformation without adaptive remeshing would lead to the loss of high frequency information.

The last major improvement of MeshGraphNets over GNS is extending the output vector $Y$ with additional components to also predict further targets, such as the stress field.

In difference to the Graph Network-based Simulators, the input here includes a predefined mesh and the output is extended to contain dynamical features like pressure.

\begin{figure}
    \centering
    \includegraphics[width=0.48\textwidth]{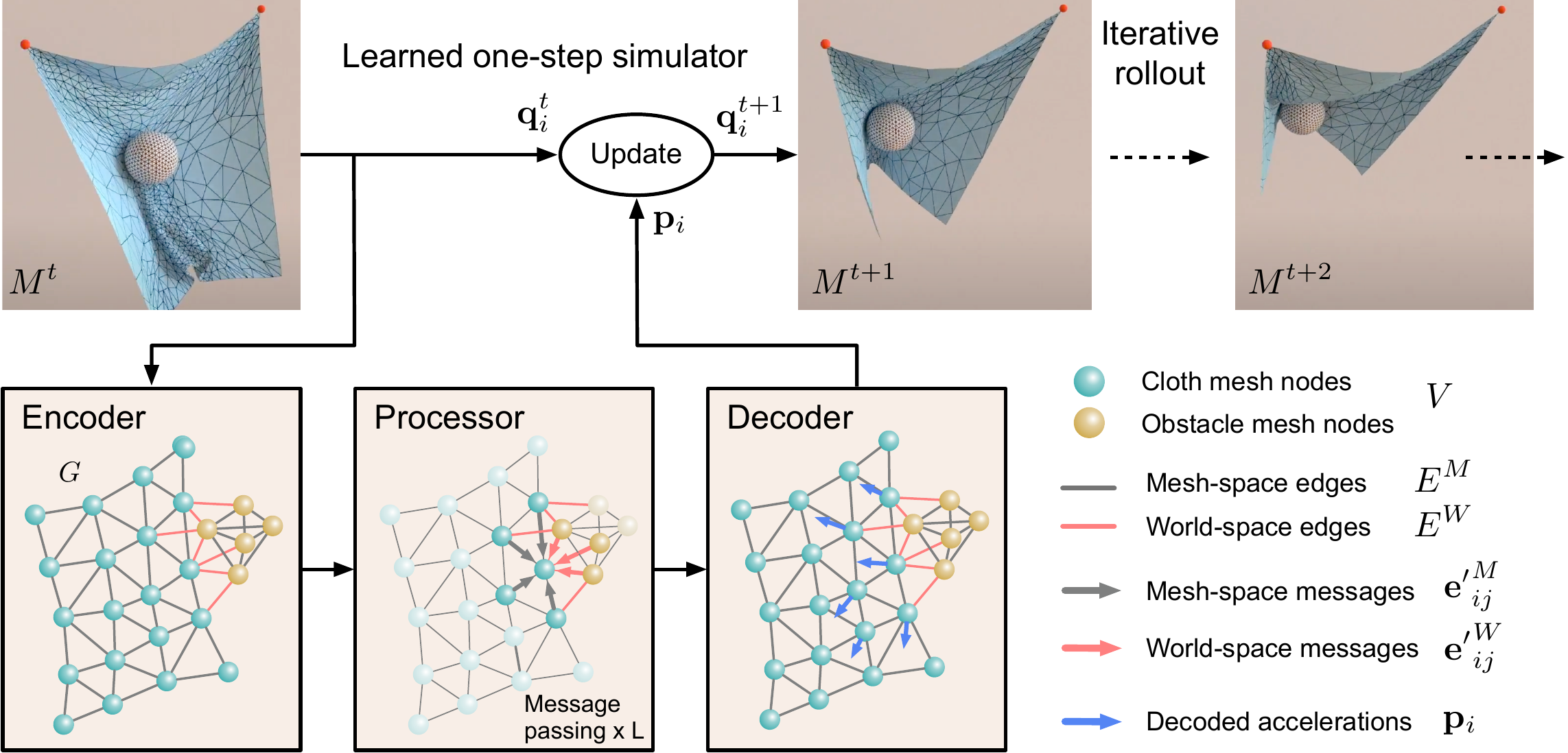}
    \caption{Illustration of the MeshGraphNets scheme with a decomposition of its algorithm into the encoder, processor, and decoder (Image source: \citet{pfaff2021learning}).}
    \label{fig:mgn_scheme}
\end{figure}

% From MD to GNS
\subsection{Similarities between the Development Trajectories of Particle-based Methods and Graph Neural Network-based Approaches to Simulations}

Beginning with Molecular Dynamics, the earliest and most fundamental particle-based method, we will now outline the similarities between the development trajectories, and the derivations inherent to them, of MeshGraphNets and the development of particle-based methods in physics.

% The root of all evil: Molecular Dynamics
\subsubsection{Similarities to Molecular Dynamics}
Molecular Dynamics is a widely used simulation method which generates the trajectories of an N-body atomic system. For the sake of intellectual clarity we restrict ourselves to its simplest form, the unconstrained Hamiltonian mechanics description.

The construction of connections, and edges is one of the clearest similarities between Molecular Dynamics and MeshGraphNets. Both can potentially have a mesh as an input, and both compute the interactions based on spatial distances up to a fixed threshold. Iterative updates, or the repeated application of Graph Network layers in the MeshGraphNets, extend the effective interaction radius beyond the immediate neighbourhood of a particle such that all particles can be interacted with. Both approaches are at the same time translationally invariant w.r.t. accelerations, and permutation equivariant w.r.t. the particles, and use a symplectic time-integrator. While there are theoretical reasons for this choice in Molecular Dynamics, it is choice of convenience in the context of learned approaches. The main difference between the two approaches lies in the computation of the accelerations. In Molecular Dynamics the derivative of a predefined potential function is evaluated, whereas a learned model is used in the Graph Network-based Simulators.

% SPH
\subsubsection{Similarities to Smoothed Particle Hydrodynamics}
A closer relative to the Graph Network-based Simulators is the Smoothed Particle Hydrodynamics algorithm originating from astrophysics~\cite{lucy1977numerical,gingold1977smoothed}. Smoothed Particle Hydrodynamics discretizes the governing equations of fluid dynamics, the Navier-Stokes equations, with kernels such that the discrete particles follow Newtonian mechanics with the equivalent of a prescribed molecular potential. Both, Smoothed Particle Hydrodynamics, and Graph Network-based Simulators obey the continuum assumption, whereas Molecular Dynamics presumes a discrete particle distribution, and is constrained to extremely short time intervals.

% Comparison
\subsubsection{The Differences}
Summarizing the key differences between the closely related approaches, Molecular Dynamics and Smoothed Particle Hydrodynamics both take one past state $X^t$ as an input, whereas Graph-based approaches require a history of $k$ states ${\bf{X}}^{t_{0:K}}$. Molecular Dynamics encodes geometric relations in the potential, MeshGraphNets encode the geometry in the mesh, while there exists no direct way for inclusion in the other two approaches. Molecular Dynamics, and Smoothed Particle Hydrodynamics explicitly encode physical laws, for learned methods all these parameters and relations have to be learned from data.

% Connecting the FPM
A key advancement of MeshGraphNets, coming from the Graph Network-based Simulators, is the explicit superimposition of solutions on both sets of edges, which far outperforms the implicit distinction of interactions. This approach is equally applicable to all conventional particle-, and mesh-based simulations in engineering. Borrowing the Fluid Particle Model from fluid mechanics, we can subsequently connect the classical methods with the learned approaches by viewing meshes and particles as the same entity under the fluid-particle paradigm.

\subsubsection{Connecting MeshGraphNets to Graph Neural Network-based Simulations with the Fluid Particle Model}
\begin{figure}
    \centering
    \includegraphics[width=0.45\textwidth]{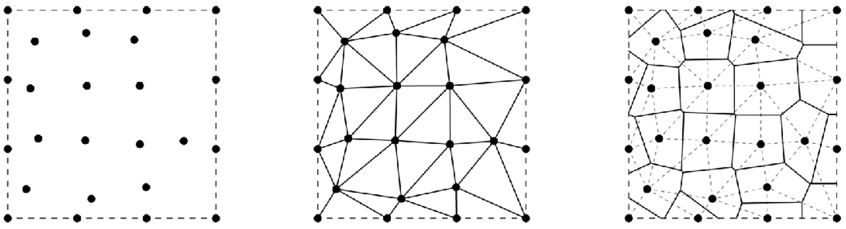}
    \caption{Single points (left), Delaunay triangulation (middle), and Voronoi diagram (right)(Image source: \citet{rokicki2016voronoi}}
    \label{fig:voronoi_tess}
\end{figure}

The Fluid Particle Model~\cite{espanol1998fluid} is a mesoscopic Newtonian model, as seen in Figure~\ref{fig:sec2_overview}, situated on an intermediate scale between the microscopic Molecular Dynamics and the macroscopic Smoothed Particle Hydrodynamics. It views particles from the point of view of a Voronoi tesselation of the molecular fluid, see Figure~\ref{fig:voronoi_tess}. The Voronoi tesselation coarse-grains the atomistic system to a pseudoparticle system with ensembles of atoms in thermal equilibrium summarized as pseudoparticles. This pseudoparticle construction is closely related to the MeshGraphNets construction, where each mesh node also corresponds to the cell center of a simulated pseudoparticle. Smoothed Particle Hydrodynamics as well as Dissipative Particle Dynamics~\cite{hoogerbrugge1992simulating} also both operate on pseudoparticles. All of these approaches share that they have to presume a large enough number of atoms per pseudoparticles to be viewed as a thermodynamic system. 

% Similarity to MGN
Especially in Dissipative Particle Dynamics one injects Gaussian noise to approximate a physical system, just as is done for Graph Network-based Simulators and MeshGraphNets to stabilize the training. We surmise that this injection of noise into graph-based simulators amounts to forcing the learned model to predict the true output despite the noisy inputs, hence leading the model to converge to the central limit of the estimated conditional distribution of the acceleration.

The construction of Voronoi tesselations governs that the size of the cells is to be inversely proportional to variations in their properties, hence leading to more sampling in regions with high property variation. The very same argument based on the curvature as a heuristic is being used to derive the mesh refinement of the MeshGraphNets algorithm.

\section{Relation to Numerical Schemes}

After the recent success of Neural ODEs solvers \cite{chen2018neuralode}, it has taken almost four years to start considering Neural PDEs in general \cite{brandstetter2022message}. By definition, PDEs deal with derivatives of multiple variables, compared to ODEs having one variable. As a result, typical numerical approximations of PDEs are much more diverse depending on the peculiarities of the PDE of interest. Typical PDE solvers operating on grids (Eulerian description) include Finite Difference Methods (FDM), Finite Volume Methods (FVM), and Finite Element Methods (FEM), whereas other methods follow the trajectory of irregularly spaced points (Lagrangian description) like Smoothed Particle Hydrodynamics (SPH), Fluid Particle Model (FPM), Dissipative Particle Dynamics (DPD) \cite{hoogerbrugge1992simulating}, Volume of Fluid Method (VOF) \cite{hirt1981volume}, Particle-in-Cell (PIC) \cite{brackbill1986flip}, Material Point Method (MPM) \cite{sulsky1993particle}, Discrete Element Method (DEM) \cite{cundall1979discrete}, and Meshless FEM (MFEM). Finally, there are also approaches to solving PDEs without any discretization as in \citet{sawhneyseyb22gridfree}. Each of these methods works best for a specific type of PDE, boundary/initial conditions, and parameter range. In this section we compare concepts from these classical methods to state-of-the-art learned algorithms.

\subsection{Data augmentation with white noise}
Two popular papers corrupting training inputs with additive Gaussian noise include \citet{sanchez2020learning,pfaff2021learning}, as described before. The goal of this approach is to force the model to deal with accumulating noise leading to a distribution shift during longer rollouts. Thus, the noise acts as an effective regularization technique, which in these two papers allows for much longer trajectories than seen during training. However, one major issue with this approach is that the scale of the noise is represented by two new hyperparameters, which have to be tuned manually (\citet{pfaff2021learning}, Appendix 2.2).

A perspective on noise injection coming from the physical sciences is to see it through the lens of mesoscopic particle methods like the Fluid Particle Model and Dissipative Particle Dynamics, in which the noise originates from the Brownian motion at small scales. Although GNS and MeshGraphNets operate on scales too large for the relevance of Brownian motion, the Fluid Particle Model provides a principled way of relating particle size and noise scale. The underlying considerations from statistical mechanics might aid to a better understanding of the influence of training noise and in turn make approaches based on it more efficient.

\subsection{Data augmentation by multi-step loss}
Another way of dealing with the distribution shift is by training a model to correct its own mistakes via some form of a multi-step loss, i.e. during training a short trajectory is generated and the loss is summed over one or multiple past steps \cite{tompson2017accelerating,um2020solver,Ummenhofer2020Lagrangian,brandstetter2022message}. The results on this vary with some researchers reporting better performance than with noise injection~\cite{brandstetter2022message}, while others report the opposite experience~\cite{sanchez2020learning}.

Looking at classical solvers for something related to the multi-step loss, it is natural to think of adaptive time integrators used by default in ODE routines like ODE45 in Matlab \cite{dormand1980family}. Adaptive integrators work by generating two short trajectories of the same time length, but with different step sizes, and as long as the outcome with larger steps differs within some bounds, then the step size is increased. This guarantees some level of long-term rollout stability just as attempted with the multi-step loss, but the multi-step loss forces the network to implicitly correct for future deviations of the trajectory without actually changing the step size. The adaptive step-size idea has gained popularity in ML with the introduction of Neural ODEs~\cite{chen2018neuralode}. 

% \subsection{Data augmentation with Lie point symmetry}
% In a recent paper by \citet{Brandstetter2022LiePS} it was demonstrated how to use the symmetries of a PDE to augment the training data very effectively. In this manner, the difficult task of designing arbitrary Lie equivariant layers in a neural network \cite{finzi2020generalizing,macdonald2021enabling} is replaced by a much simpler preprocessing routine. It is difficult to relate this topic to conventional PDE solvers bacause solvers are by construction equivariant, which we discuss further in the following.

% Other papers on Lie groups and NN:
% https://arxiv.org/abs/2002.12880 
% https://arxiv.org/abs/1909.12057 
% https://arxiv.org/abs/2109.07103 
% https://arxiv.org/abs/2111.08251 
% https://arxiv.org/abs/2012.10885

\subsection{Equivariance bias}
Numerical PDE solvers come in two flavors: stencil-based and kernel-based, both of which are equivariant to translation, rotation, and reflection in space (Euclidean group equivariance), as well as translation in time (by Noether's theorem). These properties arise from the conservation of energy, which is a fundamental principle in physics. While equivariance, with respect to the Euclidean group, has been around for a couple of years on grids \cite{weiler20183d}, its extension to the grid-free (Lagrangian) setting is gaining popularity just recently \cite{brandstetter2021geometric,schutt2021painn,batzner20223,batzner2022allegro}. Here, we talk about equivariance in terms of a neural net operation on vectors, which rotates the output exactly the same way as the input is rotated, as opposed to working with scalar values, which is called an invariant operation, e.g. SchNet \cite{schutt2017schnet}. The performance boost by including equivariant features is significant and reaches up to an order of magnitude compared to invariant methods \cite{batzner20223}.

% The connection to particle-based methods like SPH might be confusing: for a learned method

\subsection{Input multiple past steps}
Another common performance improvement in neural net training is observed by stacking multiple past states as an input \cite{sanchez2020learning,pfaff2021learning,brandstetter2022message}. One argument supporting this approach is overfitting prevention by inputting more data \cite{pfaff2021learning}. Looking at conventional solvers we very rarely see multiple past states as input and this is done for materials with memory property, e.g. rheological fluids or "smart" materials. Thus, providing multiple past states implicitly assumes that there is some nonphysical non-Markovian retardation process, which in most cases does not correspond to the physics used for training data generated.

The only physical justification of a multi-step input we are aware of arises if we train the model to learn a coarse-grained representation of the system. \citet{li2015incorporation} showed that explicit memory effects are necessary in Dissipative Particle Dynamics for the correct coarse-graining of a complex dynamical system using the Mori-Zwanzig formalism. Given that papers like GNS and MeshGraphNets do not make use of coarse-graining, it is questionable why we observe improvement in performance and whether this trick generalizes well to different settings.

\subsection{Spatial multi-scale modeling}
Conventional multi-scale methods include, among others, all types of coarse-graining, Wavelet-based methods (e.g. \citet{kim2008wavelet}), and the Fast Multipole Method \cite{rokhlin1985rapid}. Graph Networks seem especially suitable for tasks like coarse-graining as they are designed to work on unstructured domains, opposed for example to approaches using Wavelet or Fourier transforms, which require regular grids. GNNs seem especially promising with many applications in Molecular Dynamics \cite{husic2020coarse} and engineering \cite{lino2021simulating,valencia2022remusgnn,migus2022multiscale,han2022predicting}. It is particularly interesting to see works like \citet{migus2022multiscale} inspired by multi-resolution methods and \citet{valencia2022remusgnn} resembling geometric coarse-graining by weighted averaging. All these methods rely on the fact that physical systems exhibit multi-scale behavior, meaning that the trajectory of a particle depends on its closest neighbors, but also on more far-reaching weaker forces. Splitting the scales and combining their contributions can greatly reduce computation. One of the great advantages of GNNs is their capability to operate on irregularly spaced data, which is necessary for most coarse-graining approaches.

\subsection{Locality of interactions}
In most cases, graph-based approaches to solving PDEs define the edges in the graph, based on an interaction radius. Methods using the Graph Network architecture \cite{battaglia2018relational} effectively expand the receptive field of each node with every further layer, in the extreme case resulting in the phenomenon known as over-smoothing. But if we keep the number of layers reasonably low, the receptive field will always be larger compared to a conventional simulation with the same radius. Until recently, it was thought that a large receptive field is the reason for the success of learned simulators, but \citet{batzner2022allegro} question that assumption. In this paper, an equivariant graph network with fixed interaction neighbors performs on a par with the very similar Graph Network-based method NequIP \cite{batzner20223} on molecular property prediction tasks. This finding supports the physics-based argument about the locality of interactions.

% Illustrative figure to capture the keywords for the methods in this, and the following sections.
\begin{figure*}[h!]
    \centering
    %  trim={<left> <lower> <right> <upper>}
    \includegraphics[trim={7cm 5.5cm 2cm 7cm},clip,width=0.9\textwidth]{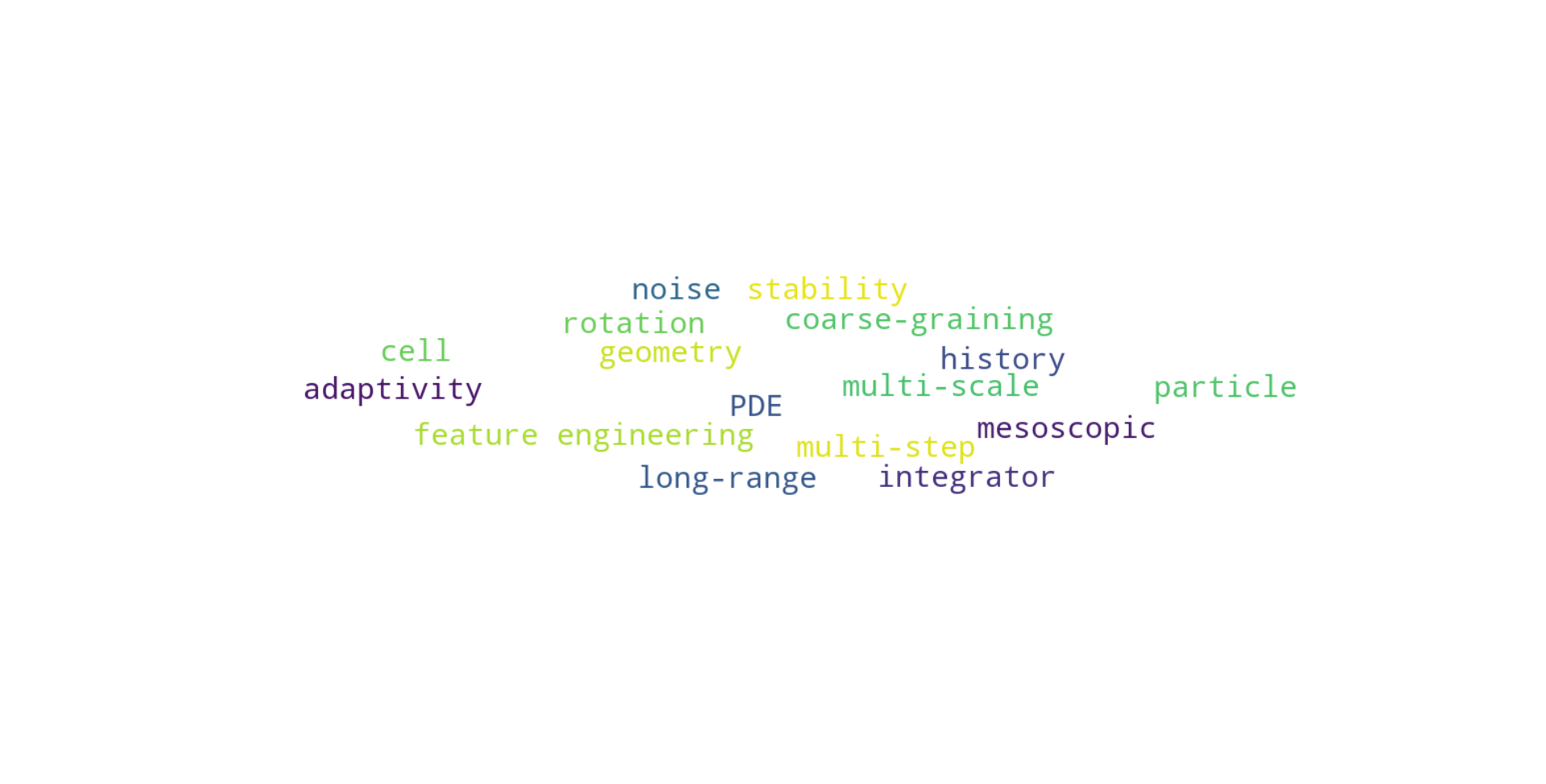}
    \vspace{-1.5cm}
    \caption{Overview of the currently under-utilized ideas discussed in Section~4 for Machine Learning approaches for the physical sciences.}
    \label{fig:wordcloud}
\end{figure*}

\subsection{Mesh vs Particle}
GNN-based simulation approaches offer the flexibility to combine particles and meshes out-of-the-box. If we then train one neural network to reproduce the results of a Finite Element solution on a mesh and Smoothed Particle Hydrodynamics solution over particles, this is where learned methods really shine. This was achieved with the MeshGraphNets framework \cite{pfaff2021learning}. We argue that the transition from particles to meshes is a direct result of a coarse-graining procedure using Voronoi tessellation, which is related to the derivation of the Fluid Particle Model. The main assumption in this derivation is that each mesh cell should be small enough that it can be treated as being in equilibrium - similar to the assumption made when discretizing a domain with points.

\subsection{Stencils}
We talk about stencils when operating on regular grids. Although this is not the main strength of GNNs, there are some useful concepts from stencil-based simulations, which are conventionally nontrivial to generalize to particles, but can easily be adapted with GNNs. \citet{brandstetter2022message} state that their paper is motivated by the observation that the Weighted Essentially Non-Oscillatory scheme (WENO) \cite{shu1998essentially} can be written as a special case of a GNN. Another work, inspired by the general idea of the Finite Volume Method, looking at the fluxes at the left and right cell boundary, was developed by \citet{praditia2021finite}.  Inspired by the Finite Element Method, finite element networks were introduced by weighting the contributions of neighbouring cells by their volume, as is done in Finite Element analysis~\cite{lienen_fen2022}.

\subsection{Integration schemes}
In addition to the time-step adaptation mentioned in relation to multi-step losses, another topic investigated in literature is the order of the integrator \cite{sanchez2019hamiltonian}. This work points to the fact that higher order integrators lead to much better robustness, with respect to the choice of an integration time step. Another interesting question discussed in this paper is whether symplectic integrators improve performance of a learned Hamiltonian neural net. The answer seems to be that the symplectic property is much less important than the order of the integrator, which is in contrast with conventional Molecular Dynamics integrators, which work extremely poorly if not symplectic.

% Sources of Inspiration from the Development of Classical Approaches
% Further Sources of Inspiration from Classical Approaches

\section{Untapped Ideas from Classical Approaches}

In this subsection, we introduce potentially useful ideas from conventional differential equation solvers in science, which to the best of our knowledge have not been adapted in main-stream learned PDE solvers yet. Figure~\ref{fig:wordcloud} is a collection of these concepts in the form of a word cloud.

\subsection{Noise during inference}
Adding noise to the inputs during training has proven to be useful, but has not been done during testing. One idea would be to use noise during inference to emulate Brownian motion. And one further topic we already mentioned is the relation of the noise scale to particle mass. From mesoscopic methods and the Fluctuation-dissipation theorem we would expect the noise to scale as $1/\sqrt{m}$ if a coarser representation is used.

\subsection{Multiple time steps}
Learned Molecular Dynamics simulations stick to using only the last past state and doing the same for larger-scale simulations might partially explain the unphysical behavior of the GNS method demonstrated in \citet{klimesch2022simulating}. For coarse-graining though a longer history might be helpful.

\subsection{Feature Engineering}
From the Volume of Fluid Method we could adapt the idea of including features corresponding to the ratio of different material, if we are interested in simulating multi-material flows. The Discrete Element Method suggests encoding much more features like rotational degree of freedom (in magnetic field or simulating friction), stateful contact information (contact simulations), and often complicated geometry (for non-spherical, e.g. granular particles). Inspired by shock-capturing methods used routinely for the solution of nonlinear fluid dynamics problems \cite{leveque2020riemann}, one could think of further hand-tuned node features indicating the presence of a shock.

\subsection{Particles and Grid}
There are a number of methods using the best of both particle and grid worlds like the Particle-in-Cell method and its successor Material Point Method. The idea of updating the node features and from time to time also based on the grid cell they belong to, might speed up simulations and is worth exploring. Now, if we restrict ourselves to regularly spaced particles, respectively grid cells, our solver toolkit becomes much richer with methods like the Fast Fourier Transform (which has already seen great success with the Fourier Neural Operator~\cite{li2020fourier}) and the Wavelet Transform (as used in the PDE-Net~\cite{long2018pde}) at our disposal, as mentioned above in the context of multi-scale modeling.

% What's missing here is the race in Neural ODEs/CDEs to design faster, and more efficient integrators -> see Patrick Kidger, Christopher Salvi

\subsection{Integrator}
Taking the perspective of Neural ODEs~\cite{chen2018neuralode} with the neural network learning the perfect acceleration, one could arguably expect the next evolutionary step to be the combination of learned integrators with adaptive integration schemes. Incorporating insights from classical numerical methods, one should possibly seek to define an equivalent stability criterion for learned methods as the Courant-Friedrichs-Lewy (CFL) condition for classical numerical methods. This would in turn aid in bounding the time, and subsequently explore time steps smaller than the critical value.

\section{Conclusion \& Discussion}

In this article, we claim that studying classical PDE solvers and their past development offers a direct path to the acceleration of the development of learned PDE solvers. Examples in literature show that biasing a learned solver by means of architectural design, data augmentation, feature engineering, etc. incorporating existing knowledge from classical solvers can greatly improve performance, explainability, and data-efficiency.

In Section~2 we show how this development has already subconsciously played out in the development of graph-based learned solvers following the same development as particle-based methods such as Molecular Dynamics, Smoothed Particle Hydrodynamics, and the Fluid-Particle Model. This investigation is revisited for algorithmic comparisons and illustrations of the limitations of classical solvers later on. In Section~3 we then focus on ideas from classical approaches which have found their way into recent learned solver literature, and discuss the physical interpretation of these developments. In the discussed examples, the included physically motivated biases are used to improve robustness w.r.t. hyperparameter choices, lower errors, and speed-up inference. 

Section~4 takes a glimpse into a possible version of the future with ideas which have, to the best of our knowledge, not yet been integrated in learned methods. Given the elaborate history of classical methods, and the short, but highly dynamic history of learned approaches, there is still a lot of potential to be realized within the latter by incorporating insights from the former.  

Going further, many exciting problems in the physical sciences, such as simulations involving multiple spatial scales, multiple temporal scales, non-Newtonian fluids, or phase-changing materials, are heavily data-constrained and will hence have to rely on insights from classical methods for Machine Learning approaches to become feasible.

% Although one of the strengths of GNNs lies in using nonhomogeneously distributed nodes, learned approaches can easily adapt strategies developed for stencil-based methods and then extend them to the nonhomogeneous setting.
% Next to improving performance on classical benchmarks like the flow around a cylinder, learned approaches have a great potential to accelerate more complex simulations involving multiple spatial and temporal scales, as well as non-Newtonian or phase-changing materials. 

% Not in line with the story at all
%Further, coarse-graining seems to be another very promising use of GNNs, in which case many ongoing improvements in graph clustering algorithms can be directly applied to physical systems. 

\bibliography{RootsGNS}

\begin{thebibliography}{51}
\providecommand{\natexlab}[1]{#1}
\providecommand{\url}[1]{\texttt{#1}}
\expandafter\ifx\csname urlstyle\endcsname\relax
  \providecommand{\doi}[1]{doi: #1}\else
  \providecommand{\doi}{doi: \begingroup \urlstyle{rm}\Url}\fi

\bibitem[Battaglia et~al.(2018)Battaglia, Hamrick, Bapst, Sanchez-Gonzalez,
  Zambaldi, Malinowski, Tacchetti, Raposo, Santoro, Faulkner,
  et~al.]{battaglia2018relational}
Battaglia, P.~W., Hamrick, J.~B., Bapst, V., Sanchez-Gonzalez, A., Zambaldi,
  V., Malinowski, M., Tacchetti, A., Raposo, D., Santoro, A., Faulkner, R.,
  et~al.
\newblock Relational inductive biases, deep learning, and graph networks.
\newblock \emph{arXiv preprint arXiv:1806.01261}, 2018.

\bibitem[Batzner et~al.(2022)Batzner, Musaelian, Sun, Geiger, Mailoa,
  Kornbluth, Molinari, Smidt, and Kozinsky]{batzner20223}
Batzner, S., Musaelian, A., Sun, L., Geiger, M., Mailoa, J.~P., Kornbluth, M.,
  Molinari, N., Smidt, T.~E., and Kozinsky, B.
\newblock E (3)-equivariant graph neural networks for data-efficient and
  accurate interatomic potentials.
\newblock \emph{Nature communications}, 13\penalty0 (1):\penalty0 1--11, 2022.

\bibitem[Brackbill \& Ruppel(1986)Brackbill and Ruppel]{brackbill1986flip}
Brackbill, J.~U. and Ruppel, H.~M.
\newblock Flip: A method for adaptively zoned, particle-in-cell calculations of
  fluid flows in two dimensions.
\newblock \emph{Journal of Computational physics}, 65\penalty0 (2):\penalty0
  314--343, 1986.

\bibitem[Brandstetter et~al.(2021)Brandstetter, Hesselink, van~der Pol,
  Bekkers, and Welling]{brandstetter2021geometric}
Brandstetter, J., Hesselink, R., van~der Pol, E., Bekkers, E., and Welling, M.
\newblock Geometric and physical quantities improve e (3) equivariant message
  passing.
\newblock \emph{arXiv preprint arXiv:2110.02905}, 2021.

\bibitem[Brandstetter et~al.(2022)Brandstetter, Worrall, and
  Welling]{brandstetter2022message}
Brandstetter, J., Worrall, D.~E., and Welling, M.
\newblock Message passing neural {PDE} solvers.
\newblock In \emph{International Conference on Learning Representations}, 2022.

\bibitem[Brunton et~al.(2016)Brunton, Proctor, and
  Kutz]{brunton2016discovering}
Brunton, S.~L., Proctor, J.~L., and Kutz, J.~N.
\newblock Discovering governing equations from data by sparse identification of
  nonlinear dynamical systems.
\newblock \emph{Proceedings of the national academy of sciences}, 113\penalty0
  (15):\penalty0 3932--3937, 2016.

\bibitem[Chen et~al.(2018)Chen, Rubanova, Bettencourt, and
  Duvenaud]{chen2018neuralode}
Chen, R. T.~Q., Rubanova, Y., Bettencourt, J., and Duvenaud, D.~K.
\newblock Neural ordinary differential equations.
\newblock In Bengio, S., Wallach, H., Larochelle, H., Grauman, K.,
  Cesa-Bianchi, N., and Garnett, R. (eds.), \emph{Advances in Neural
  Information Processing Systems}, volume~31. Curran Associates, Inc., 2018.

\bibitem[Cranmer et~al.(2020)Cranmer, Sanchez~Gonzalez, Battaglia, Xu, Cranmer,
  Spergel, and Ho]{cranmer2020discovering}
Cranmer, M., Sanchez~Gonzalez, A., Battaglia, P., Xu, R., Cranmer, K., Spergel,
  D., and Ho, S.
\newblock Discovering symbolic models from deep learning with inductive biases.
\newblock \emph{Advances in Neural Information Processing Systems},
  33:\penalty0 17429--17442, 2020.

\bibitem[Cundall \& Strack(1979)Cundall and Strack]{cundall1979discrete}
Cundall, P.~A. and Strack, O.~D.
\newblock A discrete numerical model for granular assemblies.
\newblock \emph{geotechnique}, 29\penalty0 (1):\penalty0 47--65, 1979.

\bibitem[Dormand \& Prince(1980)Dormand and Prince]{dormand1980family}
Dormand, J.~R. and Prince, P.~J.
\newblock A family of embedded runge-kutta formulae.
\newblock \emph{Journal of computational and applied mathematics}, 6\penalty0
  (1):\penalty0 19--26, 1980.

\bibitem[Espanol(1998)]{espanol1998fluid}
Espanol, P.
\newblock Fluid particle model.
\newblock \emph{Physical Review E}, 57\penalty0 (3):\penalty0 2930, 1998.

\bibitem[Frey et~al.(2022)Frey, Soklaski, Axelrod, Samsi, Gomez-Bombarelli,
  Coley, and Gadepally]{frey2022neural}
Frey, N., Soklaski, R., Axelrod, S., Samsi, S., Gomez-Bombarelli, R., Coley,
  C., and Gadepally, V.
\newblock Neural scaling of deep chemical models.
\newblock 2022.

\bibitem[Gingold \& Monaghan(1977)Gingold and Monaghan]{gingold1977smoothed}
Gingold, R.~A. and Monaghan, J.~J.
\newblock Smoothed particle hydrodynamics: theory and application to
  non-spherical stars.
\newblock \emph{Monthly notices of the royal astronomical society},
  181\penalty0 (3):\penalty0 375--389, 1977.

\bibitem[Han et~al.(2022)Han, Gao, Pfaff, Wang, and Liu]{han2022predicting}
Han, X., Gao, H., Pfaff, T., Wang, J.-X., and Liu, L.
\newblock Predicting physics in mesh-reduced space with temporal attention.
\newblock In \emph{International Conference on Learning Representations}, 2022.

\bibitem[Hirt \& Nichols(1981)Hirt and Nichols]{hirt1981volume}
Hirt, C.~W. and Nichols, B.~D.
\newblock Volume of fluid (vof) method for the dynamics of free boundaries.
\newblock \emph{Journal of computational physics}, 39\penalty0 (1):\penalty0
  201--225, 1981.

\bibitem[Hoogerbrugge \& Koelman(1992)Hoogerbrugge and
  Koelman]{hoogerbrugge1992simulating}
Hoogerbrugge, P. and Koelman, J.
\newblock Simulating microscopic hydrodynamic phenomena with dissipative
  particle dynamics.
\newblock \emph{EPL (Europhysics Letters)}, 19\penalty0 (3):\penalty0 155,
  1992.

\bibitem[Husic et~al.(2020)Husic, Charron, Lemm, Wang, P{\'e}rez, Majewski,
  Kr{\"a}mer, Chen, Olsson, de~Fabritiis, et~al.]{husic2020coarse}
Husic, B.~E., Charron, N.~E., Lemm, D., Wang, J., P{\'e}rez, A., Majewski, M.,
  Kr{\"a}mer, A., Chen, Y., Olsson, S., de~Fabritiis, G., et~al.
\newblock Coarse graining molecular dynamics with graph neural networks.
\newblock \emph{The Journal of chemical physics}, 153\penalty0 (19):\penalty0
  194101, 2020.

\bibitem[Ketcheson et~al.(2020)Ketcheson, LeVeque, and del
  Razo]{leveque2020riemann}
Ketcheson, D.~I., LeVeque, R.~J., and del Razo, M.~J.
\newblock \emph{Riemann Problems and Jupyter Solutions}.
\newblock Society for Industrial and Applied Mathematics, Philadelphia, PA,
  2020.
\newblock \doi{10.1137/1.9781611976212}.

\bibitem[Kim et~al.(2008)Kim, Th{\"u}rey, James, and Gross]{kim2008wavelet}
Kim, T., Th{\"u}rey, N., James, D., and Gross, M.
\newblock Wavelet turbulence for fluid simulation.
\newblock \emph{ACM Transactions on Graphics (TOG)}, 27\penalty0 (3):\penalty0
  1--6, 2008.

\bibitem[Klimesch et~al.(2022)Klimesch, Holl, and
  Thuerey]{klimesch2022simulating}
Klimesch, J., Holl, P., and Thuerey, N.
\newblock Simulating liquids with graph networks.
\newblock \emph{arXiv preprint arXiv:2203.07895}, 2022.

\bibitem[Kochkov et~al.(2021)Kochkov, Smith, Alieva, Wang, Brenner, and
  Hoyer]{kochkov2021machine}
Kochkov, D., Smith, J.~A., Alieva, A., Wang, Q., Brenner, M.~P., and Hoyer, S.
\newblock Machine learning--accelerated computational fluid dynamics.
\newblock \emph{Proceedings of the National Academy of Sciences}, 118\penalty0
  (21), 2021.

\bibitem[Li et~al.(2015)Li, Bian, Li, and Karniadakis]{li2015incorporation}
Li, Z., Bian, X., Li, X., and Karniadakis, G.~E.
\newblock Incorporation of memory effects in coarse-grained modeling via the
  mori-zwanzig formalism.
\newblock \emph{The Journal of chemical physics}, 143\penalty0 (24):\penalty0
  243128, 2015.

\bibitem[Li et~al.(2020)Li, Kovachki, Azizzadenesheli, Liu, Bhattacharya,
  Stuart, and Anandkumar]{li2020fourier}
Li, Z., Kovachki, N., Azizzadenesheli, K., Liu, B., Bhattacharya, K., Stuart,
  A., and Anandkumar, A.
\newblock Fourier neural operator for parametric partial differential
  equations.
\newblock \emph{arXiv preprint arXiv:2010.08895}, 2020.

\bibitem[Lienen \& G\"unnemann(2022)Lienen and G\"unnemann]{lienen_fen2022}
Lienen, M. and G\"unnemann, S.
\newblock Learning the dynamics of physical systems from sparse observations
  with finite element networks.
\newblock In \emph{International Conference on Learning Representations
  (ICLR)}, 2022.

\bibitem[Lino et~al.(2021)Lino, Cantwell, Bharath, and
  Fotiadis]{lino2021simulating}
Lino, M., Cantwell, C., Bharath, A.~A., and Fotiadis, S.
\newblock Simulating continuum mechanics with multi-scale graph neural
  networks.
\newblock \emph{arXiv preprint arXiv:2106.04900}, 2021.

\bibitem[Long et~al.(2018)Long, Lu, Ma, and Dong]{long2018pde}
Long, Z., Lu, Y., Ma, X., and Dong, B.
\newblock Pde-net: Learning pdes from data.
\newblock In \emph{International Conference on Machine Learning}, pp.\
  3208--3216. PMLR, 2018.

\bibitem[Lucy(1977)]{lucy1977numerical}
Lucy, L.~B.
\newblock A numerical approach to the testing of the fission hypothesis.
\newblock \emph{The astronomical journal}, 82:\penalty0 1013--1024, 1977.

\bibitem[Migus et~al.(2022)Migus, Yin, Mazari, and patrick
  gallinari]{migus2022multiscale}
Migus, L., Yin, Y., Mazari, J.~A., and patrick gallinari.
\newblock Multi-scale physical representations for approximating {PDE}
  solutions with graph neural operators.
\newblock In \emph{ICLR 2022 Workshop on Geometrical and Topological
  Representation Learning}, 2022.

\bibitem[Musaelian et~al.(2022{\natexlab{a}})Musaelian, Batzner, Johansson,
  Sun, Owen, Kornbluth, and Kozinsky]{batzner2022allegro}
Musaelian, A., Batzner, S., Johansson, A., Sun, L., Owen, C.~J., Kornbluth, M.,
  and Kozinsky, B.
\newblock Learning local equivariant representations for large-scale atomistic
  dynamics, 2022{\natexlab{a}}.

\bibitem[Musaelian et~al.(2022{\natexlab{b}})Musaelian, Batzner, Johansson,
  Sun, Owen, Kornbluth, and Kozinsky]{musaelian2022learning}
Musaelian, A., Batzner, S., Johansson, A., Sun, L., Owen, C.~J., Kornbluth, M.,
  and Kozinsky, B.
\newblock Learning local equivariant representations for large-scale atomistic
  dynamics.
\newblock \emph{arXiv preprint arXiv:2204.05249}, 2022{\natexlab{b}}.

\bibitem[Pfaff et~al.(2021)Pfaff, Fortunato, Sanchez-Gonzalez, and
  Battaglia]{pfaff2021learning}
Pfaff, T., Fortunato, M., Sanchez-Gonzalez, A., and Battaglia, P.
\newblock Learning mesh-based simulation with graph networks.
\newblock In \emph{International Conference on Learning Representations}, 2021.

\bibitem[Praditia et~al.(2021)Praditia, Karlbauer, Otte, Oladyshkin, Butz, and
  Nowak]{praditia2021finite}
Praditia, T., Karlbauer, M., Otte, S., Oladyshkin, S., Butz, M.~V., and Nowak,
  W.
\newblock Finite volume neural network : Modeling subsurface contaminant
  transport.
\newblock In \emph{ICLR 2021 : Ninth International Conference on Learning
  Representations}. Cornell University, 2021.
\newblock \doi{10.48550/arXiv.2104.06010}.

\bibitem[Rokhlin(1985)]{rokhlin1985rapid}
Rokhlin, V.
\newblock Rapid solution of integral equations of classical potential theory.
\newblock \emph{Journal of computational physics}, 60\penalty0 (2):\penalty0
  187--207, 1985.

\bibitem[Rokicki \& Gawell(2016)Rokicki and Gawell]{rokicki2016voronoi}
Rokicki, W. and Gawell, E.
\newblock Voronoi diagrams--architectural and structural rod structure research
  model optimization.
\newblock \emph{MAZOWSZE Studia Regionalne}, \penalty0 (19):\penalty0 155--164,
  2016.

\bibitem[Sanchez-Gonzalez et~al.(2019)Sanchez-Gonzalez, Bapst, Cranmer, and
  Battaglia]{sanchez2019hamiltonian}
Sanchez-Gonzalez, A., Bapst, V., Cranmer, K., and Battaglia, P.
\newblock Hamiltonian graph networks with ode integrators.
\newblock \emph{arXiv preprint arXiv:1909.12790}, 2019.

\bibitem[Sanchez-Gonzalez et~al.(2020)Sanchez-Gonzalez, Godwin, Pfaff, Ying,
  Leskovec, and Battaglia]{sanchez2020learning}
Sanchez-Gonzalez, A., Godwin, J., Pfaff, T., Ying, R., Leskovec, J., and
  Battaglia, P.
\newblock Learning to simulate complex physics with graph networks.
\newblock In \emph{International Conference on Machine Learning}, pp.\
  8459--8468. PMLR, 2020.

\bibitem[Sawhney et~al.(2022)Sawhney, Seyb, Jarosz, and
  Crane]{sawhneyseyb22gridfree}
Sawhney, R., Seyb, D., Jarosz, W., and Crane, K.
\newblock Grid-free {Monte} {Carlo} for {PDEs} with spatially varying
  coefficients.
\newblock \emph{ACM Transactions on Graphics (Proceedings of SIGGRAPH)},
  41\penalty0 (4), July 2022.
\newblock \doi{10.1145/3528223.3530134}.

\bibitem[Scarselli et~al.(2008)Scarselli, Gori, Tsoi, Hagenbuchner, and
  Monfardini]{scarselli2008graph}
Scarselli, F., Gori, M., Tsoi, A.~C., Hagenbuchner, M., and Monfardini, G.
\newblock The graph neural network model.
\newblock \emph{IEEE transactions on neural networks}, 20\penalty0
  (1):\penalty0 61--80, 2008.

\bibitem[Sch{\"u}tt et~al.(2017)Sch{\"u}tt, Kindermans, Sauceda~Felix, Chmiela,
  Tkatchenko, and M{\"u}ller]{schutt2017schnet}
Sch{\"u}tt, K., Kindermans, P.-J., Sauceda~Felix, H.~E., Chmiela, S.,
  Tkatchenko, A., and M{\"u}ller, K.-R.
\newblock Schnet: A continuous-filter convolutional neural network for modeling
  quantum interactions.
\newblock \emph{Advances in neural information processing systems}, 30, 2017.

\bibitem[Sch{\"u}tt et~al.(2021)Sch{\"u}tt, Unke, and
  Gastegger]{schutt2021painn}
Sch{\"u}tt, K., Unke, O., and Gastegger, M.
\newblock Equivariant message passing for the prediction of tensorial
  properties and molecular spectra.
\newblock In Meila, M. and Zhang, T. (eds.), \emph{Proceedings of the 38th
  International Conference on Machine Learning}, volume 139 of
  \emph{Proceedings of Machine Learning Research}, pp.\  9377--9388. PMLR,
  18--24 Jul 2021.

\bibitem[Shu(1998)]{shu1998essentially}
Shu, C.-W.
\newblock Essentially non-oscillatory and weighted essentially non-oscillatory
  schemes for hyperbolic conservation laws.
\newblock In \emph{Advanced numerical approximation of nonlinear hyperbolic
  equations}, pp.\  325--432. Springer, 1998.

\bibitem[Sulsky et~al.(1993)Sulsky, Chen, and Schreyer]{sulsky1993particle}
Sulsky, D., Chen, Z., and Schreyer, H.~L.
\newblock A particle method for history-dependent materials.
\newblock Technical report, Sandia National Labs., Albuquerque, NM (United
  States), 1993.

\bibitem[Tompson et~al.(2017)Tompson, Schlachter, Sprechmann, and
  Perlin]{tompson2017accelerating}
Tompson, J., Schlachter, K., Sprechmann, P., and Perlin, K.
\newblock Accelerating {E}ulerian fluid simulation with convolutional networks.
\newblock In Precup, D. and Teh, Y.~W. (eds.), \emph{Proceedings of the 34th
  International Conference on Machine Learning}, volume~70 of \emph{Proceedings
  of Machine Learning Research}, pp.\  3424--3433. PMLR, 06--11 Aug 2017.

\bibitem[Um et~al.(2020)Um, Brand, Fei, Holl, and Thuerey]{um2020solver}
Um, K., Brand, R., Fei, Y.~R., Holl, P., and Thuerey, N.
\newblock Solver-in-the-loop: Learning from differentiable physics to interact
  with iterative pde-solvers.
\newblock In Larochelle, H., Ranzato, M., Hadsell, R., Balcan, M., and Lin, H.
  (eds.), \emph{Advances in Neural Information Processing Systems}, volume~33,
  pp.\  6111--6122. Curran Associates, Inc., 2020.

\bibitem[Ummenhofer et~al.(2020)Ummenhofer, Prantl, Thuerey, and
  Koltun]{Ummenhofer2020Lagrangian}
Ummenhofer, B., Prantl, L., Thuerey, N., and Koltun, V.
\newblock Lagrangian fluid simulation with continuous convolutions.
\newblock In \emph{International Conference on Learning Representations}, 2020.

\bibitem[Valencia et~al.(2022)Valencia, Fotiadis, Bharath, and
  Cantwell]{valencia2022remusgnn}
Valencia, M.~L., Fotiadis, S., Bharath, A.~A., and Cantwell, C.~D.
\newblock {REM}us-{GNN}: A rotation-equivariant model for simulating continuum
  dynamics.
\newblock In \emph{ICLR 2022 Workshop on Geometrical and Topological
  Representation Learning}, 2022.

\bibitem[Veličković et~al.(2020)Veličković, Ying, Padovano, Hadsell, and
  Blundell]{Velickovic2020Neural}
Veličković, P., Ying, R., Padovano, M., Hadsell, R., and Blundell, C.
\newblock Neural execution of graph algorithms.
\newblock In \emph{International Conference on Learning Representations}, 2020.

\bibitem[Wang et~al.(2020)Wang, Kashinath, Mustafa, Albert, and
  Yu]{wang2020towards}
Wang, R., Kashinath, K., Mustafa, M., Albert, A., and Yu, R.
\newblock Towards physics-informed deep learning for turbulent flow prediction.
\newblock In \emph{Proceedings of the 26th ACM SIGKDD International Conference
  on Knowledge Discovery \& Data Mining}, pp.\  1457--1466, 2020.

\bibitem[Wang et~al.(2022)Wang, Xu, Cai, Miller, Smidt, Wang, Tang, and
  G{\'o}mez-Bombarelli]{wang2022generative}
Wang, W., Xu, M., Cai, C., Miller, B.~K., Smidt, T., Wang, Y., Tang, J., and
  G{\'o}mez-Bombarelli, R.
\newblock Generative coarse-graining of molecular conformations.
\newblock \emph{arXiv preprint arXiv:2201.12176}, 2022.

\bibitem[Weiler et~al.(2018)Weiler, Geiger, Welling, Boomsma, and
  Cohen]{weiler20183d}
Weiler, M., Geiger, M., Welling, M., Boomsma, W., and Cohen, T.~S.
\newblock 3d steerable cnns: Learning rotationally equivariant features in
  volumetric data.
\newblock \emph{Advances in Neural Information Processing Systems}, 31, 2018.

\bibitem[Xu et~al.(2020)Xu, Li, Zhang, Du, ichi Kawarabayashi, and
  Jegelka]{xu2020what}
Xu, K., Li, J., Zhang, M., Du, S.~S., ichi Kawarabayashi, K., and Jegelka, S.
\newblock What can neural networks reason about?
\newblock In \emph{ICLR}, 2020.

\end{thebibliography}
\bibliographystyle{icml2022}

\end{document}